\documentclass[10pt, a4paper]{article}
\usepackage{lrec2006}
\usepackage{latexsym}

\usepackage{listings}
\title{The ALVIS Format for Linguistically Annotated Documents}

\name{A. Nazarenko, E. Alphonse, J. Derivi\`ere, T. Hamon, G. Vauvert, D. Weissenbacher} 
\address{Laboratoire  dÕInformatique de Paris-Nord (UMR 7030)\\
               University Paris 13 \& CNRS \\ 
               99, Av. J.-B. Cl\'ement, 93430 Villetaneuse, France \\
               \{firstname.name\}@lipn.univ-paris13.fr}
\abstract{The paper describes the ALVIS annotation format designed for the indexing of large collections of documents in topic-specific search engines. This paper is exemplified on the biological domain and on MedLine abstracts, as developing a specialized search engine for biologists is one of the ALVIS case studies. The ALVIS principle for linguistic annotations is based on existing works and standard propositions. We made the choice of stand-off annotations rather than inserted mark-up. Annotations are encoded as XML elements which form the linguistic subsection of the document record. }

\begin{document}

\maketitleabstract

\section{Introduction}

One of the objectives of the ALVIS project\footnote{ALVIS is a FP6 STREP projet aiming at developing an open source prototype of a distributed semantic-based search engine. See http://www.alvis.info.} is to develop semantic-based search engines that achieve good performance in information retrieval (IR) in specialized domains. As one of our case study, we are developing  a search engine for the biological domain able to answer complex queries (boolean and even relational queries including normalized gene names, for instance). In this context, we are experimentally studying the contribution of Natural Language Processing (NLP) in IR. We are testing various indexing methods based on various types of linguistic annotation. 

This paper presents and discusses the format that has been adopted in the ALVIS project for the linguistic annotation of the documents. It shows how NLP tools can add new annotations to a given document or exploit existing ones. The paper focuses on the linguistic part of the document annotations, disregarding the metadata associated with the document at the crawling step \cite{buntine&al05}.

Section~2 presents the ALVIS project and explains the need for linguistically annotated documents. The ALVIS format for linguistic annotation is presented in section~3. Section~5 discusses the advantages and limits of our format.

\section{Context: the ALVIS project}

The ALVIS project aims at building a peer-to-peer network of semantic search engines and at developing open source components to help designing new topic-specific search engines. Among these components, there is a NLP line, which goal is to enrich crawled documents with linguistic annotations allowing a semantic and domain specific indexing of these documents. The type and quality of the annotations vary with the following factors:
\begin{itemize}
\item The way the annotated documents are used: for each specific topic, beside the collection indexing, we exploit a sample of documents for acquiring specialized linguistic resources (acquisition phase). The resulting resources (named entity dictionaries, terminologies, semantic tags) are then used to tune the generic NLP line for the corresponding specific domain (production phase). A deep analysis of the documents is required at the acquisition level whereas indexing calls for an efficient and shallow analysis strategy.
\item The availability of domain specific resources: the more domain specific knowledge is available (or acquired), the richer the document annotations can be. 
\item The language of the documents: in ALVIS, four different languages are processed (English, French, Slovene, Chinese) but the various processing steps are not equally important for all languages (e.g. traditional word segmentation is useless for Chinese, lemmatisation is more important for Slovene than for English and even for French);
\item The volume of textual data to process: since NLP is computationally expensive, the deepness of document analysis depends on the efficiency of the NLP components and the size of the collection to analyze.
\end{itemize}

One of the objectives of the ALVIS project is to test the various combinations of annotations to identify which ones have a significant impact on IR results, within a given specific domain. In this context, the definition of a format for the linguistic annotation of documents is a critical issue. It must be compatible with the modularity of the ALVIS NLP line and the interchangeability of NLP tools.

\section{ALVIS format for linguistic annotations}

The linguistic annotation is represented as a layered set of textual units and linguistic properties.

\subsection{Annotation principle}

The ALVIS principle for linguistic annotations \cite{nazarenko&al04} is based on existing works and standard propositions \cite{grishman97,bird&liberman99}. We made the choice of stand-off annotations rather than inserted mark-up and annotations are encoded as XML elements which form the linguistic subsection of the document record \cite{buntine&al05}. The principle of stand-off annotations has numerous advantages:
\begin{itemize}
\item The initial textual data may be read-only and/or very large, so
  copying it to insert mark-up may be unacceptable.
\item The distribution of the initial data may be controlled whereas
  the mark-up is intended to be freely available.
\item The stand-off annotations do not pollute the initial
  textual data.
\item Stand-off annotations allow embedded and overlapping annotations that are
  incompatible with an inserted mark-up. It is easier to encode concurrent annotations
 produced by different NLP tools, non linear linguistic entities (such as  "{\em to} completely {\em decide}") and relations between
  elements belonging to various levels in the hierarchy of
  annotations.
\item New levels of annotations can be added without disturbing the
  existing ones.
\item Editing one level of annotation has minimal knock-on effects on
  others.
\item Each level of annotation can be stored and handled separately,
  eventually in several files.
\end{itemize}
The main drawback of the standoff annotation principle is that it is
difficult and computationally expensive to rebuild the textual signal
from the list of annotations.

The problem of linguistic annotation representation has
been widely studied since the beginning of the nineties and several formats have been proposed \cite{grishman97,bird&liberman99}.  The efforts to unify these
formats in order to allow interoperability among NLP tools are recent. An ISO proposition (TC37SC4/TEI) is currently under definition \cite{ide&al04},
    which will include a Feature Structure Representation, a
    Morpho-Syntactic Annotation Framework, a Category Data Repository,
    a Linguistic Annotation Framework, a Lexical Mark-up Framework and
    some Data Category S-Electronic Lexical Resources.

Our goal is not as general as that of the TC37SC4/TEI: strictly complying with the norm would make our annotation formalism more complex whereas a light version is sufficient for ALVIS needs. 


\subsection{Textual entities}

 In ALVIS, we
consider five different levels of textual units. At a basic level, the text is
segmented into tokens and the other units (words, phrases, semantic units and sentences) are built on this first level. 
The ALVIS format for these textual units is homogeneous from one level
to another. Except for tokens, each textual unit has an identifier, a
list of components and an optional form in which the sequence of
characters to which it corresponds can be copied.

For sake of readability, the following examples are given with
traditional inserted annotation (slash of brackets). The actual ALVIS format is shown on Figure~\ref{fig:sampl-output-PF}.

\subsubsection{Tokens}
\label{sec:tokens}

Tokens are the fundamental textual units in the ALVIS text processing
line. This segmentation, which is not linguistically grounded, is used for reference purpose. This level of annotation is recommended by the
TC37SC4/TEI workgroup, even if we refer to the character offset to
mark the token boundaries, rather than inserting
pointer mark-up in the textual signal.
To simplify further processing, we distinguish different types of tokens:
\begin{itemize}
\item Alphabetical tokens: sequences of letters (\texttt{a-z} and \texttt{A-Z}) including accented characters;
\item Numerical tokens: sequences of digits (\texttt{0-9});
\item Separating tokens: sequence of separator characters (\texttt{space},
  \texttt{return} ...);
\item  Symbolic tokens: any other character.
\end{itemize}

The tokenisation is the basic stage of text analysis. Tokens are
numbered from 1 for the first token. All others annotations refer
directly or indirectly to that token numbering. 
In the example of figure~\ref{t}, the 
slashes represent the token boundaries (note that blanks are
tokens).

\begin{figure}[h]
\begin{center}
\fbox{\parbox{8cm}{
/Transcription/ /of/ /the/ /cotB/,/ 
/cotC/,/ /and/ /cotX/ /genes/ /by/ 
/final/ /sigma/(/K/)/ /RNA/ 
/polymerase/ /is/ /activated/ 
/by/ /a/ /small/,/ /DNA/-/binding/ 
/protein/ /called/ /GerE/./
}}
\caption{Tokenization.}
\label{t}
\end{center}
\end{figure}

\subsubsection{Words}
\label{sec:words}
Words are the basic linguistic units. Every
word is made of one or several tokens, numeric, alphabetic or
symbolic (note that words may contain spaces, i.e.  \textit{pomme de terre} in French), even if some character strings are not trivially split into words. For example "doesn't" is made of the words "does" and "not", which do
not appear as such and which are
created independently of the corresponding tokens ({\em doesn}, the
apostrophe (') and {\em t}).

In the following example (fig.~\ref{w}), words are delimited by square brackets.
Note that neither the punctuation marks nor the blanks are words and that 
the word segmentation does not split the eventual pre-identified named entities (i.e. sigma(K)), which are annotated as semantic units (see the section~\ref{sec:semantic-units} below).

\begin{figure}[h]
\begin{center}
\fbox{\parbox{8cm}{
[Transcription] [of] [the] [cotB], [cotC], 
[and] [cotX] [genes] [by] [final] [sigma(K)] 
[RNA] [polymerase] [is] [activated] [by] 
[a] [small], [DNA-binding] [protein] 
[called] [GerE].
}}
\caption{Word segmentation.}
\label{w}
\end{center}
\end{figure}

\subsubsection{Phrases}
\label{sec:phrases}
A phrase is a group of words (or a single word) that form a
syntactic unit. It is composed of a head 
and of optional modifier(s) that can be words or phrases. 
The phrasal level only describes the phrase frontiers (see fig.~\ref{p}), not their
syntactic categories (see the section~\ref{sec:morpho-synt-prop} below). 

\begin{figure}[h]
\begin{center}
\fbox{\parbox{8cm}{
[During [sporulation of Bacillus 
subtilis]], [[spore coat proteins] 
[encoded [by [cot genes]]]] [are 
expressed [in [the mother cell]] and 
[deposited [on [the forespore]]]]. 
Transcription [of [the cotB, cotC, and 
cotX genes]] [by [final [sigma(K) RNA 
polymerase]]] [is activated [by [a small, 
[DNA-binding protein] [called GerE]]]].
[The promoter region [of [each [of [these 
genes]]]]] [has [two [GerE binding sites]]].
}}
\caption{Phrase identification.}
\label{p}
\end{center}
\end{figure}

\subsubsection{Semantic units}
\label{sec:semantic-units}
The semantic units are the textual units that are considered as significant from a semantic point of view. They can be:
\begin{itemize}
\item Named entities that refer to well identified domain entities
  (esp. proper names).
\item Terms:expressions referring to the concepts
  specific to the domain of the text.
\item Undefined semantic units: other types of relevant semantic units can be
  identified, even if their semantic status is not established.
\end{itemize}
%

In the example of the figure~\ref{su}, the named entities and terms are tagged as
XML-like inserted mark-up.

\begin{figure}[h]
\begin{center}
\fbox{\parbox{8cm}{
During sporulation of $<$NE$>$Bacillus 
subtilis$<$NE$>$, $<$term$>$spore coat 
proteins$<$term$>$ encoded by $<$term$>$
$<$NE$>$cot$<$NE$>$ genes$<$term$>$ are 
expressed in the $<$term$>$mother 
cell$<$term$>$ and deposited on the 
$<$term$>$forespore$<$term$>$. Transcription 
of the $<$NE$>$cotB$<$NE$>$, $<$NE$>$cotC$<$NE $>$, 
and $<$NE$>$cotX$<$NE$>$genes by final 
$<$NE$>$sigma(K)$<$NE$>$...
}}
\caption{Named entity and term tagging.}
\label{su}
\end{center}
\end{figure}
In our NLP platform, the entity recognition is launched before word and sentence segmentation in order to help identify certain strings which would hinder further linguistic analysis if they were not identified as semantic units. It also associates semantic tags to the identified semantic units.

\subsubsection{Sentences}
\label{sec:sentences}
The sentences correspond to a traditional textual unit. They usually start from a word with a capital initial character and ends with a period. However various other types of sentences can be encountered in texts. In the ALVIS linguistic
annotation format, we consider that titles, some list items and captions are sentences.
\begin{figure}[h]
\begin{center}
\fbox{\parbox{8cm}{
[During sporulation of Bacillus subtilis, spore 
coat proteins encoded by cot genes are 
expressed in the mother cell and deposited 
on the forespore.]
[Transcription of the cotB, cotC, and cotX 
genes by final sigma(K) RNA polymerase is 
activated by a small, DNA-binding protein 
called GerE.]
[The promoter region of each of these genes 
has two GerE binding sites.]
}}
\caption{Sentence segmentation.}
\end{center}
\end{figure}




\subsection{Properties of textual entities}
\label{sec:morpho-synt-prop}
Various properties can be associated with textual units. They are encoded as separate XML entities referring to the textual entities to which they are associated: 
\begin{itemize}
\item Morpho-syntactic tags: the lemmas, stems, syntactic categories\footnote{A syntactic category is either a phrasal category, such as noun
  phrase or verb phrase or a simple part of
  speech category such as noun or verb, which cannot be
  further
  decomposed.}, and
morpho-syntactic features.
\item Syntactic relations, which define the role (or function) played by two
words between one another. These relations are represented as triplets:
a relation type T, its head H  and its expansion E (or
modifier).
\item Semantic tags:  the semantic types, which are attached to semantic units (either named entities, terms or undefined semantic units).
\item Semantic relations : the anaphoric and domain specific relations are attached to semantic units corresponding to words or phrases.
\end{itemize}

  \begin{figure*}[htbp]
    \centering
{\scriptsize
\begin{lstlisting}[breaklines]
<documentCollection>
<documentRecord id="A79ACA58DEB7E6114747710B9A85059F">
  <acquisition>
      <acquisitionData>
        <modifiedDate>2004-11-21 15:59:14</modifiedDate>
        <urls>
          <url>http://www.ncbi.nlm.nih.gov/entrez/query.fcgi?cmd=Retrieve&amp;db=pubmed&amp;dopt=MEDLINE&amp;list_uids=10788508</url>
        <urls>
      <acquisitionData>
      <canonicalDocument>        
        <section>
          <section title="Combined action of two transcription factors regulates genes encoding spore coat proteins of Bacillus subtilis.">
	    <section>Combined action of two transcription factors regulates genes encoding spore coat proteins of Bacillus subtilis.
	    </section>
            ...
      </canonicalDocument>
  </acquisition>
\end{lstlisting}
\begin{minipage}{0.48\linewidth}
\begin{lstlisting}[breaklines]
  <linguisticAnalysis>
    <token_level>
    <token>
      <content>Combined</content>
      <from>0</from>
      <id>token1</id>
      <to>7</to>
      <type>alpha</type>
    </token>
    ...
    </token_level>
    <sentence_level>
    <sentence>
      <form>Combined action of two transcription factors regulates genes encoding spore coat proteins of Bacillus subtilis .</form>
      <id>sentence1</id>
      <refid_end_token>token30</refid_end_token>
      <refid_start_token>token1</refid_start_token>
    </sentence>
    ...
    </sentence_level>
    <semantic_unit_level>
    <semantic_unit>
      <named_entity>
        <form>Bacillus subtilis</form>
        <id>named_entity0</id>
        <list_refid_token>
            <refid_token>token27</refid_token>
            <refid_token>token28</refid_token>
            <refid_token>token29</refid_token>
         </list_refid_token>
      </named_entity>
    </semantic_unit>
    ...
    </semantic_unit_level>
    <word_level>
    <word>
      <form>Combined</form>
      <id>word1</id>
      <list_refid_token>
          <refid_token>token1</refid_token>
      </list_refid_token>
    </word>
    ...
    </word_level>
  \end{lstlisting}
   \end{minipage}
  \begin{minipage}{0.50\linewidth}
\begin{lstlisting}[breaklines]
    <lemma_level>
    <lemma>
      <canonical_form>combined</canonical_form>
      <id>lemma1</id>
      <refid_word>word1</refid_word>
    </lemma>
    ...
    </lemma_level>
    <morphosyntactic_features_level>
    <morphosyntactic_features>
      <id>morphosyntactic_features1</id>
      <refid_word>word1</refid_word>
      <syntactic_category>JJ</syntactic_category>
    </morphosyntactic_features>
    <morphosyntactic_features>
      <id>morphosyntactic_features10</id>
      <refid_word>word10</refid_word>
      <syntactic_category>NN</syntactic_category>
    </morphosyntactic_features>
    ...
    </morphosyntactic_features_level>
  <syntactic_relation_level>
    <syntactic_relation>
      <id>syntrel1</id>
      <syntactic_relation_type>NCOMPby
      </syntactic_relation_type>
      <refid_head>
	<refid_word>word26</refid_word>
      </refid_head>
      <refid_modifier>
	<refid_word>word35</refid_word>
      </refid_modifier>
    </syntactic_relation>
    ...
  </syntactic_relation_level>
   ...
  <semantic_features_level>
    <semantic_features>
      <id>sf5</id>
      <semantic_category>
	<list_refid_ontology_node>
	  <refid_ontology_node>species </refid_ontology_node>
	</list_refid_ontology_node>
      </semantic_category>
      <refid_semantic_unit> named_entity0 </refid_semantic_unit>
    </semantic_features>
   ...
  </linguisticAnalysis>
 </documentRecord>
</documentCollection>
  \end{lstlisting}
   \end{minipage}
}
        \caption{Example of the input and output of the linguistic annotation process.}
    \label{fig:sampl-output-PF}
  \end{figure*}

\section{Conclusion} 

In this paper, we presented the format that has been adopted to encode the linguistic annotations of documents in the ALVIS project. Besides incrementality and separability, we argue that our format meets the requirements of openness, explicitness and consistency that any linguistic annotation framework is supposed to fulfil, according to \cite{ide&al04}. 

The counterpart of explicitness is the huge size of the annotaed documents. On a collection of 55~000 web pages (biological collection comprising more than 80 millions of words), we observe that the full linguistic annotation exploiting large term and named entity dictionaries increases the size of the collection by a factor of 16. 

Fortunately, some of the linguistic annotations are only useful for NLP and can be erased at the indexing level since there are not directly useful for searching documents. In the ALVIS project, we are currently studying which NLP annotations really enhance the indexing of documents in a semantic search engine. 

%




\bibliographystyle{lrec2006}
\bibliography{lrec-nazarenko} 

\end{document}